\documentclass [times, review, 10pt]{elsarticle}
\usepackage{booktabs} 
\usepackage{graphicx} 
\usepackage{float} 

\usepackage{amssymb}
\usepackage{amsmath}

\journal{Pattern Recognition}

\begin{document}

\begin{frontmatter}


\cortext[cor1]{Corresponding author}

\title{VSE-MOT: Multi-Object Tracking in Low-Quality Video Scenes Guided by Visual Semantic Enhancement}


\author[label1]{Jun Du
} 
\ead{jun.du@bjtu.edu.cn}
\author[label1]{
Weiwei Xing\corref{cor1}
} 
\ead{wwxing@bjtu.edu.cn}
\author[label2]{
Ming Li\corref{cor1}
} 
\ead{liming@gml.ac.cn}
\author[label2]{
Fei Richard Yu
} 
\ead{yufei@gml.ac.cn}
\affiliation[label1]{organization={School of Software
Engineering},
            addressline={Beijing Jiaotong University}, 
            city={Beijing},
            postcode={100044},
            country={China}}
\affiliation[label2]{organization={Guangdong Laboratory of Artificial Intelligence and Digital Economy},
            city={Shenzhen},
            postcode={518083},
            country={China}}

\begin{abstract}
Current multi-object tracking (MOT) algorithms typically overlook issues inherent in low-quality videos, leading to significant degradation in tracking performance when confronted with real-world image deterioration. Therefore, advancing the application of MOT algorithms in real-world low-quality video scenarios represents a critical and meaningful endeavor. To address the challenges posed by low-quality scenarios, inspired by vision-language models, this paper proposes a Visual Semantic Enhancement-guided Multi-Object Tracking framework (VSE-MOT). Specifically, we first design a tri-branch architecture that leverages a vision-language model to extract global visual semantic information from images and fuse it with query vectors. Subsequently, to further enhance the utilization of visual semantic information, we introduce the Multi-Object Tracking Adapter (MOT-Adapter) and the Visual Semantic Fusion Module (VSFM). The MOT-Adapter adapts the extracted global visual semantic information to suit multi-object tracking tasks, while the VSFM improves the efficacy of feature fusion. Through extensive experiments, we validate the effectiveness and superiority of the proposed method in real-world low-quality video scenarios. Its tracking performance metrics outperform those of existing methods by approximately 8\% to 20\%, while maintaining robust performance in conventional scenarios.\vspace{\baselineskip}\vspace{\baselineskip}
\end{abstract}




\end{frontmatter}



\section{Introduction}
\label{sec1}

Multi-Object Tracking (MOT) aims to locate and associate multiple objects within video sequences. It is extensively utilized in a variety of downstream applications, such as video behavior analysis \cite{1,2}, autonomous driving \cite{3,4}, and surveillance \cite{5,6}. Recently, MOT has garnered significant attention across diverse practical scenes, greatly propelling its development. However, these endeavors are chiefly tailored for high-quality inputs and overlook the low-quality video scenes \cite{7} that are prevalent in real-world settings. Based on this, we investigate multi-object tracking in low-quality video scenes encountered in the real world.

Low-quality videos, such as those captured under conditions of low illumination, high noise, or blurriness, pose a formidable challenge to the performance of multi-object tracking (MOT) algorithms. These videos often contain incomplete information, which leads to decreased accuracy in target detection and feature extraction. Consequently, traditional tracking algorithms \cite{11,54,55,56} that rely on appearance and motion struggle to work effectively.

\begin{figure}
    \centering
    \includegraphics[width=1\linewidth]{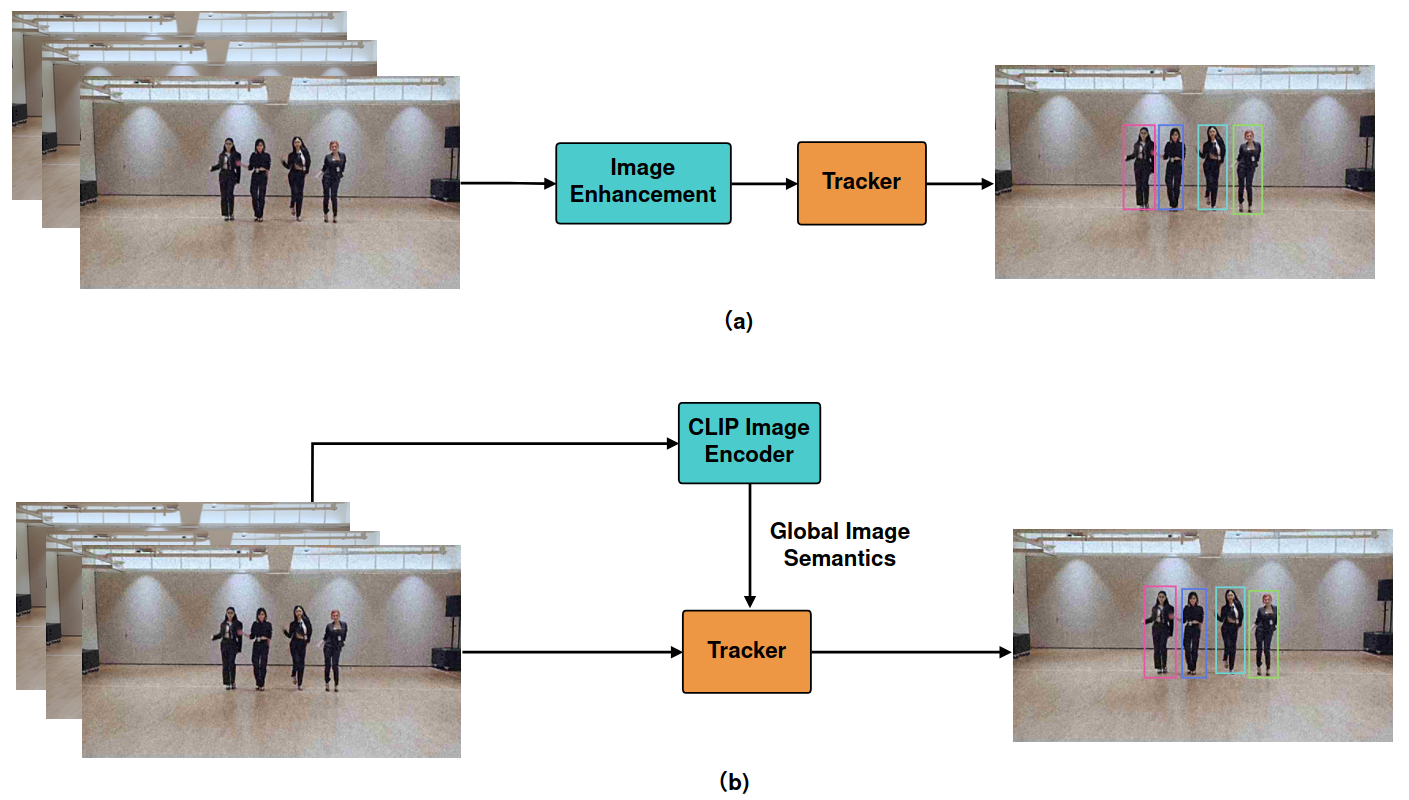}
    \caption{Comparison between (a) image enhancement tracking pipeline and (b) our proposed visual semantic enhancement tracking pipeline. Image enhancement tracking often fall short when dealing with real-world
complexity and cannot meet the tracking demands in variable
environments. visual semantic enhancement tracking learns invariant visual semantic
information under noisy and degraded quality conditions, leading to improved tracking performance.}
    \label{fig1}
\end{figure}

In existing literature, research on MOT for low-quality videos is relatively scarce and often relies on simplified degradation models or assumptions specific to certain scenes, as depicted in Figure \ref{fig1}(a). In the real-world scenarios, the task of multi-object tracking is confronted with a multitude of complexities. These include diverse types of noise, non-uniform illumination conditions, intricate backgrounds and occlusions, the diversity and dynamism of targets, varying sensor quality and resolution, changeable environmental conditions, as well as the interactions and occlusions among targets. These complexities pose significant challenges to simplified degradation models and assumptions tailored for specific scenarios, often resulting in suboptimal performance of algorithms based on such assumptions when applied in practical situations.Our method leverages the CLIP Image Encoder to extract visual semantic information, thereby compensating for lost features and effectively addressing the complexities of real-world scenarios in the task of multi-object tracking. The robustness, visual semantic comprehension, and multimodal nature of CLIP not only enhance the feature representation of the targets but also augment the algorithm's adaptability to noise, illumination variations, complex backgrounds, target diversity, disparities in sensor quality, and environmental changes.

To enhance the performance of multi-object tracking algorithms on low-quality videos, our core concept is to acquire visual semantic information that remains constant despite being in environments with noise and diminished quality. Inspired by the CLIP \cite{12} model's ability to effectively understand visual semantic information during pre-training, a Multi-Object Tracking framework guided by visual semantic enhancement(VSE-MOT) is proposed, as depicted in Figure \ref{fig1}(b). Firstly, we adopt a tri-branch architecture aimed at deeply extracting and analyzing the overall visual semantic content in images and achieving effective integration with the query vector. Secondly, to adapt the global visual semantic information extracted by the frozen CLIP image encoder \cite{12} for multi-object tracking tasks, we introduce a feature adapter called the Multi-Object Tracking Adapter (MOT-Adapter) on the visual branch of CLIP, fine-tuning it for this purpose. Lastly, to optimize the integration process of the query vector with the image's global visual semantic information, we designed and proposed a novel module called the Visual Semantic Fusion Module (VSFM), specifically for feature fusion. Extensive experiments have demonstrated the effectiveness and superiority of our proposed method in real-world low-quality video scenarios, while maintaining good performance in conventional scenarios.

In summary, our main contributions are as follows:
\begin{itemize}
     \item We propose a novel multi-object tracking method (VSE-MOT) that employs a tri-branch architecture for extracting global visual semantic information from images and fusing it with the query vector.
     \item To further enhance the utilization of visual semantic information, we introduce the MOT-Adapter and VSFM.
     \item We provide a comprehensive analysis of the proposed method. Experimental results validate the effectiveness and superiority of our method in real-world low-quality video scenarios, while maintaining good performance in conventional scenarios.
\end{itemize}

\section{Related Works}
\label{sec2}
Deep learning has been widely applied in various scenarios such as intelligent transportation \cite{li2021self,li2021exploiting,Yan_2025_CVPR}, cross-modal learning \cite{Liu_2025_CVPR, zhao2025favchat, li2025uni}, privacy protection \cite{li2023dr,li2023stprivacy}, and artificial intelligence generated content \cite{li2024instant3d, liu2024realera}. Similarly, multi-object tracking (MOT), a critical task in computer vision that aims to continuously estimate the trajectories of multiple target objects (e.g., pedestrians, vehicles, and animals) across consecutive video frames while maintaining their unique identities, has also extensively leveraged deep learning techniques.

\subsection{Detection-based Tracking Methods}
\label{subsec1}
Detection-based methods dominate the field of Multiple Object Tracking (MOT), typically following two main steps: target detection first, followed by inter-frame target association. The effectiveness of these methods largely depends on the accuracy of the initial detection phase.Various techniques attempt to utilize the Hungarian algorithm for target association. For instance, the SORT \cite{8} algorithm combines the Kalman filter \cite{15} for state prediction of targets and matches targets by calculating the Intersection over Union (IoU) between predicted and detected boxes. DeepSORT \cite{16} introduces an appearance feature extraction network on this basis, optimizing the matching process by calculating cosine distances. TrackRCNN \cite{18} further explore the joint optimization of detection and appearance features. SiamMOTION \cite{17} achieves real-time multi-object tracking through a proposal engine with an attention mechanism, an inertia-driven region-of-interest extractor, and a comparison head. SeaTrack \cite{20} constructs a tracking pipeline based on pure motion information, and achieves robust tracking through guiding modulation, multi-factor modeling, and motion fusion mechanisms. ByteTrack \cite{21} employs an advanced detector based on YOLOX \cite{22} and incorporates low-confidence detection results into the association process by improving the SORT algorithm. BoT-SORT \cite{23} enhances performance by optimizing Kalman filter parameters, compensating for camera motion, and integrating ReID features. TransMOT \cite{9} and GTR \cite{24} utilize spatio-temporal transformers to process instance features and aggregate historical information to form an assignment matrix. OC-SORT \cite{25} abandons the strict linear motion assumption in favor of a learnable dynamic model.

\subsection{Tracking by Query Propagation}
\label{subsec2}
Another paradigm of MOT extends query-based object detectors \cite{26,27,28} to tracking. These methods track the same instance by maintaining the consistency of queries among different frames. The interaction between queries and image features can be carried out in parallel or serially.
Parallel methods process short video clips by having a set of queries interact with features from all frames to predict trajectories. VisTR \cite{10} apply DETR \cite{26} to trajectory detection in short video clips. This method consume a large amount of memory due to the need to process entire videos and can only handle short video segments. 
Sequential methods update queries frame by frame, iteratively optimizing tracking results through interaction with image features. TrackFormer \cite{33} and MOTR \cite{34} extend Deformable DETR \cite{35} to predict bounding boxes and update queries for continuous tracking. MeMOT \cite{36} constructs short-term and long-term feature memory banks to generate tracking queries. TransTrack \cite{37} predicts the location of targets in the next frame through single-query propagation. MOTRv2 \cite{11} utilizes YOLOX as the object detector to generate high-quality target proposals, which are input as anchors into the improved MOTR tracker, thereby learning tracking associations. MOTIP \cite{39} regards the object association task as a contextual ID prediction problem.

\subsection{Vision-language Models}
\label{subsec3}
Vision-language Models have been extensively studied in fields such as text-to-image retrieval, visual question answering, and referential segmentation. Recently, visual-linguistic pretraining has garnered increasing attention, with the milestone work being Contrastive Language-Image Pre-training (CLIP). CLIP pre-trains models through contrastive learning among 400 million image-text pairs crawled from the Internet. It has demonstrated impressive generalization capabilities across evaluations on 30 classification datasets. The pre-trained CLIP encoders have also been applied to many other downstream tasks, such as open-vocabulary detection and zero-shot visual semantic segmentation. Recently, some follow-up works attempted to leverage the pre-trained models for video domains. For example, CLIP4Clip \cite{14} transferred the knowledge of CLIP model to the video-text retrieval, Actionclip \cite{40} attempted to capitalize on the capabilities of CLIP for the purpose of video recognition, furthermore, CLIP has been employed to address the intricate challenge of video action localization (Nag et al.) \cite{41}.

To summarize the above tracking methods,they lack consideration for low-quality video scenarios. Characteristics such as blur, noise, and uneven lighting in low-quality videos make it difficult for traditional tracking algorithms to work effectively. Our approach utilizes the rich visual semantic information from the CLIP model to further classify and associate objects in low-quality videos. This paper extracts global visual semantic information from low-quality images using the frozen CLIP Image Encoder.

\section{Method}
\label{sec3}
\begin{figure}[H]
    \centering
    \includegraphics[width=1\linewidth]{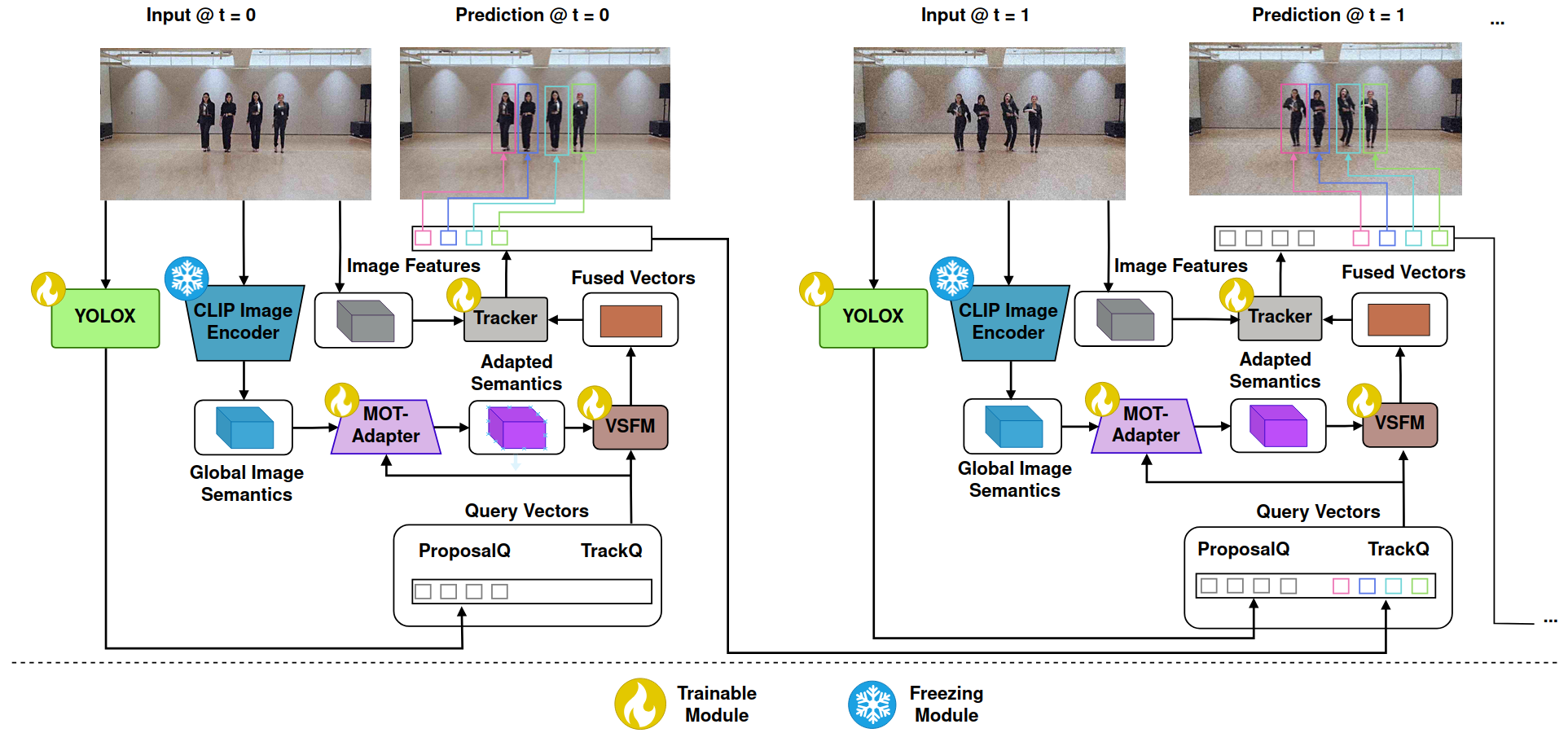}
    \caption{The overall architecture of VSE-MOT. The frozen CLIP Image Encoder extracts global visual semantic information from images, and proposals generated by the  detector YOLOX are used to produce proposal queries. Tracking inquiries are transferred from the preceding frame for the purpose of forecasting the bounding boxes of tracked objects. The combination of proposal queries and track queries generates query vectors. The global visual semantic information of the images is fused with the query vectors through the MOT-Adapter and VSFM modules to generate fused vectors. Both the fused vectors and the image features are fed into the tracker to produce predictions frame by frame.}
    \label{fig2}
\end{figure}

In this work, we adopt the MOTRv2 model as the baseline model. We propose a multi-object tracking framework named VSE-MOT, the overall framework of which is illustrated in Figure \ref{fig2}. 

Within the proposed framework, firstly, we employ a three-branch architecture that utilizes the frozen CLIP Image Encoder(ResNet50 architecture) to fully extract the global visual semantic information of images, integrating it with query vectors. The query vectors are composed of a combination of proposal queries and track queries. Specifically, proposal queries are generated based on the object proposals produced by the YOLOX detector, providing initial candidate information for potential objects in the current frame. In contrast, track queries are propagated from the previous frame and serve to predict the bounding boxes of tracked objects, thereby maintaining the temporal continuity of object tracking. Secondly, the MOT-Adapter, through spatial and channel attention mechanisms \cite{42}, extracts the spatial and channel features of the query vectors and the global visual semantic information of the image, performs multiplication and weighting on the channel dimension and spatial dimension respectively, and then adds the weighted channel features and spatial features together for integration. This enables the global visual semantic information of images extracted by the frozen CLIP Image Encoder to be adapted for multi-object tracking tasks. Lastly, the VSFM feature fusion module utilizes multi-head self-attention mechanism \cite{43} and atrous spatial pyramid pooling \cite{44} for multi-scale feature extraction and fusion, thereby better integrating the query vectors with the global visual semantic information of images.

\subsection{Multi-Object Tracking Adapter}
\label{subsec1}
\begin{figure}
    \centering
    \includegraphics[width=1\linewidth]{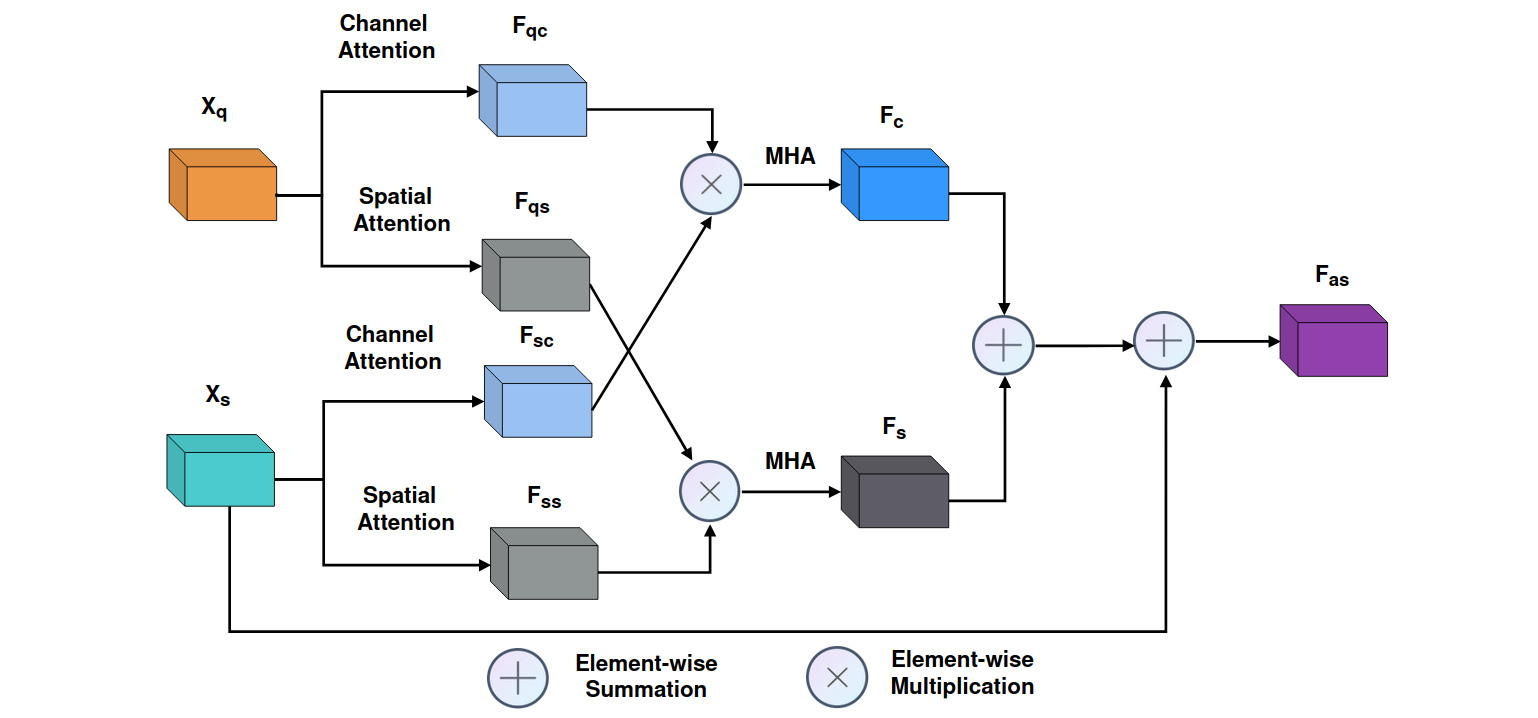}
    \caption{Structure of Multi-Object Tracking Adapter. Simple fusion lacks understanding of spatial and channel dimensions, so this module employ channel attention mechanism and spatial attention mechanism to weight and integrate the
query vector features and the global visual semantic information
features.}
    \label{fig3}
\end{figure}

The global visual semantic information extracted by the frozen CLIP Image Encoder often fails to adapt to specific downstream tasks. To enhance the adaptability of the global visual semantic information for multi-object tracking tasks, we propose a module named MOT-Adapter. Figure \ref{fig3} illustrates the structure of the proposed MOT-Adapter module. \( X_q \in \mathbb{R}^{C\times H \times W} \) and \( X_s \in \mathbb{R}^{C\times H \times W} \) represent the query vector features and the global visual semantic information features extracted by the frozen CLIP Image Encoder, respectively, where \textit{H} denotes the
height, \textit{W} denotes the width, and \textit{C} denotes the channel dimension. We believe that simple fusion lacks understanding of spatial and channel dimensions, so we employ channel attention mechanism and spatial attention mechanism to weight and integrate the query vector features and the global visual semantic information features. Initially, the feature maps \( X_{q} \) and \( X_{s} \) are each passed through identical channel attention modules \( X_{q} \), and spatial attention modules to obtain feature maps \( F_{qc} \in \mathbb{R}^{C\times H \times W} \), \( F_{qs} \in \mathbb{R}^{C\times H \times W} \), \( F_{sc} \in \mathbb{R}^{C\times H \times W} \), and \( F_{ss} \in \mathbb{R}^{C\times H \times W} \), represented as:
\begin{equation}
\label{deqn_ex1a}
{F}_{qc},{F}_{qs},{F}_{sc},{F}_{ss}={C}{A}{S}{A}({X}_{q},{X}_{s})
\end{equation}
where \textit{CASA}(·) represents the channel attention module and the spatial attention module. Next, the channel features \( F_{qc} \) and \( F_{sc} \) are multiplied and weighted, while the spatial features \( F_{qs} \) and \( F_{ss} \) are also multiplied and weighted. Then, the weighted features are passed through multi-head self-attention layers to obtain the feature maps \( F_{c} \in \mathbb{R}^{C\times H \times W} \) and \( F_{s} \in \mathbb{R}^{C\times H \times W} \), represented as:
\begin{equation}
\label{deqn_ex1a}
{F}_{c}=M HA({F}_{qc}\times {F}_{sc})
\end{equation}
\begin{equation}
\label{deqn_ex1a}
{F}_{s}=M HA({F}_{qs}\times {F}_{ss})
\end{equation}
where $ \times $ represents element-wise multiplication, and \textit{MHA}(·) represents the multi-head self-attention layer.

To enhance model performance and enable the network to learn more complex feature representations, the feature maps \( F_{c} \) and \( F_{s} \) are first added together to obtain the fused vector features. Subsequently, the global visual semantic information features \( X_{s} \)are added to the fused vector features to yield the feature map \( F_{as} \in \mathbb{R}^{C\times H \times W} \), represented as:
\begin{equation}
\label{deqn_ex1a}
F_{as}=\mathrm{X}_{s}+(F_{c}+F_{s})
\end{equation}

\subsection{Visual Semantic Fusion Module}
\label{subsec2}
\begin{figure}[H]
    \centering
    \includegraphics[width=1\linewidth]{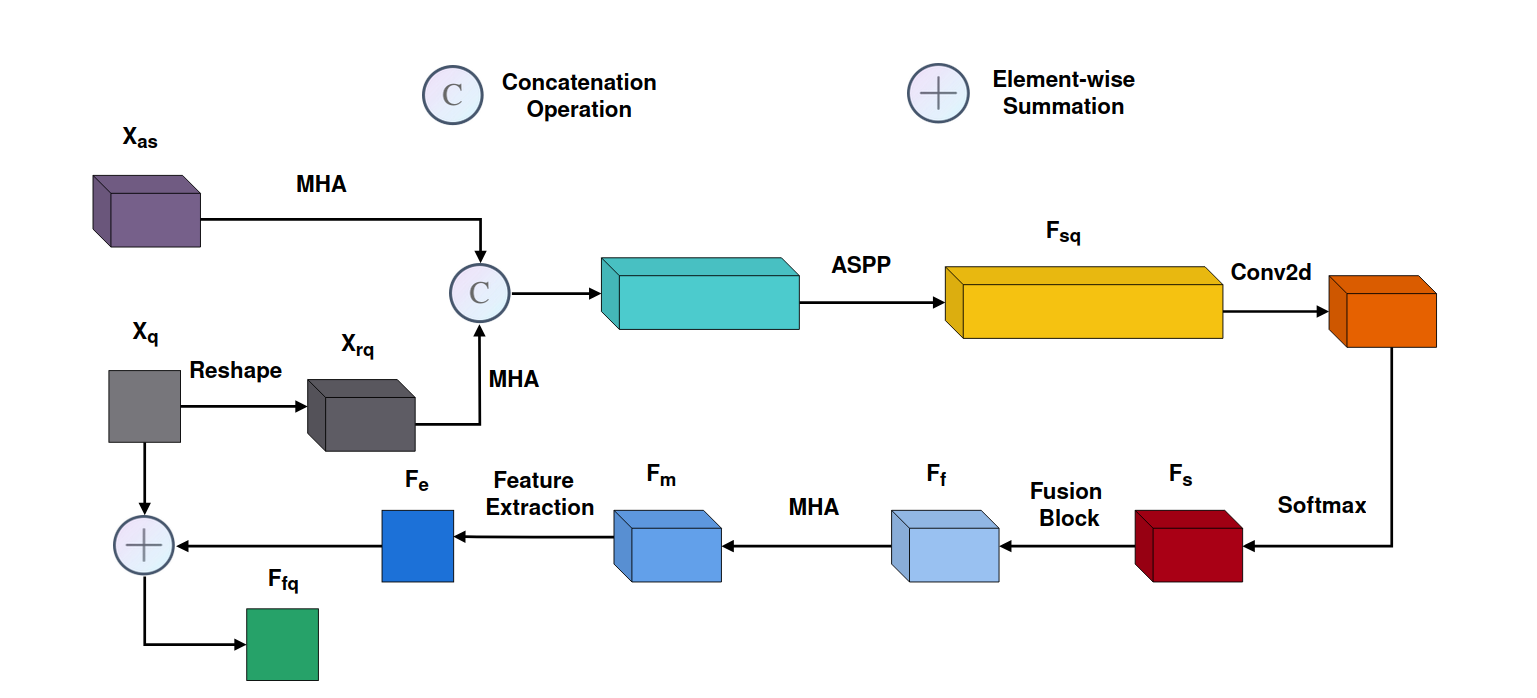}
    \caption{Structure of Visual Semantic Fusion Module. Since the query vectors and global visual semantic information of images reside in different feature spaces, the use of simple addition or concatenation fusion can lead to information conflict. This module utilizes multi-head self-attention mechanism and atrous spatial pyramid pooling for multi-scale feature extraction and fusion, thereby better integrating the query vectors with the global visual semantic information of images.}
    \label{fig4}
\end{figure}

Since the query vectors and global visual semantic information of images reside in different feature spaces, the use of simple addition or concatenation fusion can lead to information conflict. To better integrate the query vectors and global visual semantic information of images, we propose a feature fusion module named the Visual Semantic Fusion Module (VSFM). Figure \ref{fig4} illustrates the structure of the proposed VSFM module. \( X_{as} \in \mathbb{R}^{C\times H \times W} \) represents the visual semantic information features adjusted by the MOT-Adapter module. \( X_{q} \in \mathbb{R}^{N\times M } \) represents the query vector features, where \textit{N} denotes the number of queries and \textit{M} represents the dimensionality of the query vectors. To enhance the richness of representation and extract multi-scale features, the feature maps \( X_{as}  \) and \( X_{q}  \) are first processed through identical multi-head self-attention layers to obtain feature maps of the same dimensions. Subsequently, the feature maps are concatenated using a Concat operation, followed by atrous spatial pyramid pooling to achieve the fused feature \( F_{sq} \in \mathbb{R}^{C'\times H \times W} \), represented as:
\begin{equation}
\label{deqn_ex1a}
F_{sq}=Aspp(c a t(M H A(X_{as}),M H A(Reshape(X_{q}))))
\end{equation}
where \textit{MHA}(·) represents the multi-head self-attention layer, \textit{cat}(·) represents the concatenation operation, and \textit{Aspp}(·) signifies the atrous spatial pyramid pooling. Subsequently, \( F_{sq}\) is processed through a 1x1 2D convolution operation to further integrate features, followed by a Softmax activation function to obtain the feature map \( F_{s} \in \mathbb{R}^{C\times H \times W} \), represented as:
\begin{equation}
\label{deqn_ex1a}
{F}_{s}=\,S o f{t }m a x(f_{2d}(F_{sq}))
\end{equation}
where {f$_2$$_d$}(·) represents the 1x1 2D convolution operation, and \textit{Softmax}(·) denotes the Softmax activation function.
To more efficiently integrate the query vector and the global visual semantic information of images, we propose a fusion formula, denoted as \textit{f$_m$}, represented as:
\begin{equation}
\label{deqn_ex1a}
f_{m}={w}\times X_{as}+(1-w)\times X_{rq}
\end{equation}
where \( X_{as}  \) represents the visual semantic information features adjusted by the MOT-Adapter module, \( X_{rq} \in \mathbb{R}^{C\times H \times W} \) represents the query vector features, $ \times $ represents element-wise multiplication, \( w \in \mathbb{R}^{C\times H \times W} \) represents the input variable for \textit{f$_m$}, which is learned by the network. \( F_{s}\) obtains the feature map \( F_{f} \in \mathbb{R}^{C\times H \times W} \) through the fusion formula \textit{f$_m$}, which is represented as:
\begin{equation}
\label{deqn_ex1a}
F_{f}=f_{m}(F_{s})
\end{equation}

To enhance model performance and enable the network to learn more complex feature representations, we add the query vector to the fused vector. Initially, \( F_{f}\) is passed through a multi-head self-attention layer to obtain the feature map \( F_{m} \in \mathbb{R}^{C\times H \times W} \). Subsequently, the feature map \( F_{m}\) is processed through a feature extraction network to obtain a two-dimensional feature \( F_{e} \in \mathbb{R}^{N\times M } \) with the same shape as \( X_{q}\), represented as:
\begin{equation}
\label{deqn_ex1a}
F_{e}=f_{ex}(MHA(F_{\mathrm{f}}))
\end{equation}
where \textit{MHA}(·) denotes the multi-head self-attention layer, and \textit{f$_e$$_x$}(·) indicates a feature extraction network with three convolutional layers and one fully connected layer. Finally, the query vector \( X_{q}\) is added to the fused vector \( F_{e}\) to obtain the final fused vector \( F_{fq} \in \mathbb{R}^{N\times M } \), represented as:
\begin{equation}
\label{deqn_ex1a}
F_{fq}=X_{q}+F_{e}
\end{equation}

\section{Experiments}
\label{sec4}
To validate the superiority of the proposed VSE-MOT in low-quality video scenarios and its capability to maintain robust performance in traditional settings, we constructed low-quality multi-object tracking datasets: 1) LQDanceTrack based on the DanceTrack dataset, 2) LQMOT based on the MOT17 and MOT20 datasets. This section presents the experimental results of the proposed VSE-MOT and compares it with other state-of-the-art methods on the validation sets of the DanceTrack dataset, LQDanceTrack dataset, MOT dataset, and LQMOT dataset. Additionally, to thoroughly test and verify the performance of the proposed modules, we designed comprehensive ablation experiments. The experimental results demonstrate that the proposed method is effective and superior in real-world low-quality video scenarios while retaining strong performance in traditional settings.

\subsection{Dataset}
\label{subsec1}
{\bf{DanceTrack:}} DanceTrack \cite{13} is a large-scale multi-person tracking dataset designed for dance scenarios. It features uniform appearance characteristics and diverse motion patterns, which pose challenges for instance association across frames. The DanceTrack dataset contains 100 video sequences: 40 for training, 25 for validation, and 35 for testing. The average duration of these videos is 52.9 seconds. 

{\bf{MOT17:}} MOT17 \cite{61} is a commonly adopted dataset that includes 7 sequences for training and another 7 for testing. It primarily features relatively crowded street scenes where pedestrians exhibit simple and linear movement patterns.

{\bf{MOT20:}} The MOT20 \cite{62} dataset comprises 8 video sequences, with 4 allocated to the training set and 4 to the test set, focusing on high-density pedestrian tracking in complex environments.

{\bf{LQDanceTrack:}} To advance the development of the multi-object tracking field under low-quality video datasets, inspired by current works that simulate real-world degraded scenarios, we constructed a low-quality dataset called LQDanceTrack that simulates real-world low-quality video scenarios based on the DanceTrack dataset using the degradation function of Real-ESRGAN \cite{57}. Specifically, we selected 2/3 of each video sequence from the DanceTrack training set and degraded them using the Real-ESRGAN degradation function, which is expressed as:
\begin{equation}
\label{deqn_ex1a}
x=D_{n}(y)=(D_{n}\circ{\bf\cdot}\cdot{\bf\cdot}\circ D_{2}\circ D_{1})(y)
\end{equation}
where \textit{D$_1$},\textit{D$_2$},…,\textit{D$_n$ }are a series of degradation operators that successively act on the original image \textit{y}, where each \textit{D$_i$ }represents a specific degradation process, such as blurring, downsampling, noise addition, etc. $\circ $ denotes the composition operation of functions, that is, executing one function first and then using the result as the input for the next function, \textit{D$_n$}(\textit{y}) represents the output after all degradation operations. The image degradation process is expressed as:
\begin{equation}
\label{deqn_ex1a}
x=D(y)=\left[(y * k) \downarrow_r + n\right]_{J P E G}
\end{equation}
where \textit{y $*$ k} represents the convolution of the original image \textit{y} with the blur kernel \textit{k}, ↓$_r$ indicates that the blurred image undergoes downsampling to reduce resolution, with the downsampling ratio factor \textit{r}, implying that the size of the image is reduced by a factor of \textit{r} in each dimension. \textit{n} represents noise, which may include Gaussian noise, Poisson noise, etc., simulating the random errors that may be introduced during image capture or transmission. \textit{JPEG} indicates the application of \textit{JPEG} compression to the image. \textit{JPEG} is a common lossy image compression method used to reduce the size of image files, but it may further degrade image quality. In the construction process of our low-quality dataset LQDanceTrack, we set the size range of the blur kernel to (7, 21), and the types of blur kernels include isotropic blur kernels, anisotropic blur kernels, generalized isotropic blur kernels, generalized anisotropic blur kernels, plateau isotropic blur kernels, and plateau anisotropic blur kernels. We randomly select the size of the blur kernel and choose the type of blur kernel with probabilities [0.45, 0.25, 0.12, 0.03, 0.12, 0.03]. We randomly scale the image, with a scaling range of (0.15, 1.5), randomly add Gaussian and Poisson noise, and then process the image through \textit{JPEG }compression with a quality factor range of [20, 40] to obtain the output of a single image degradation process. We cascade this degradation process twice to obtain the final degraded image. At the same time, we apply the above degradation method to the DanceTrack validation set to obtain the validation set of the low-quality dataset LQDanceTrack.

{\bf{MOT, LQMOT:}} We selected 5 video sequences from the MOT17 training set and 2 video sequences from the MOT20 training set to form the MOT training set, while the remaining parts of the MOT17 and MOT20 training sets constituted the MOT validation set. Following the same construction methodology as applied to the LQDanceTrack training and validation sets, we processed the MOT training and validation sets to obtain the LQMOT training and validation sets.

{\bf{Mixed Training Sets:}} To enable the multi-object tracking model to adapt to both low-quality video scenarios in real-world environments and conventional video scenarios, we respectively combine the LQDanceTrack training set with the undegraded DanceTrack training set, and the LQMOT training set with the undegraded MOT training set, thus forming two mixed training sets where the ratio of low-quality videos to high-quality videos is 2:1. We train VSE-MOT and other state-of-the-art methods on these two mixed training sets respectively, and conduct tests on the LQDanceTrack validation set, DanceTrack validation set, LQMOT validation set, and MOT validation set separately. This is aimed at verifying the effectiveness and superiority of VSE-MOT in real-world low-quality video scenarios, as well as whether it can maintain favorable performance in conventional scenarios. Additionally, ablation experiments are carried out on the LQDanceTrack validation set and DanceTrack validation set.

\subsection{Evaluation Metrics}
We employ Higher Order Tracking Accuracy (HOTA) \cite{45} as an evaluation metric to assess our method and decompose its contributions into Detection Accuracy (DetA) and Association Accuracy (AssA). We also list Multi-Object Tracking Accuracy (MOTA) \cite{46} and Identity Filter Precision (IDF1) \cite{47} as additional evaluation metrics.

\subsection{Experimental Details}
All experiments were conducted using the Python programming language and the PyTorch framework. The training phase was performed on a device equipped with 8 NVIDIA GeForce RTX 4090 GPUs. The inference phase was executed on a device with a single NVIDIA GeForce RTX 4090 GPU. We employed the YOLOX detector to generate object proposals. To maximize proposal recall, we retained all YOLOX prediction boxes with confidence scores above 0.05 as proposals. For the mixed training set of LQDanceTrack and DanceTrack, the YOLOX detector was trained for 20 epochs on 8 GPUs. For the mixed training set of LQMOT and MOT, the YOLOX detector was trained for 50 epochs on 8 GPUs. We utilized ResNet50 \cite{48} as the backbone network for feature extraction. The entire network was trained on 8 GPUs with a batch size of 1 per GPU. For the mixed training set of LQDanceTrack and DanceTrack, following the practice of YOLOX, we adopted HSV augmentation in the training of VSE-MOT. We propagated tracking queries with confidence scores above 0.5, which naturally generates false positive (FPs; high-score but no instance, such as missing trajectories) and false negative (FNs; undetected instances) tracking queries to enhance the handling of FPs and FNs during inference. Both in the ablation studies and comparisons with state-of-the-art models, we trained for 10 epochs with a fixed segment size of 5. The sampling step within a segment was randomly selected between 1 and 10. The initial learning rate was set to $2 \times 10^{-4}$ and reduced by a factor of 10 at the \(5^{th}\) epoch. For the mixed training set of LQMOT and MOT, the number of training epochs was adjusted to 60, with the learning rate decayed at the \(40^{th}\) epoch. To improve detection performance, we also used a large number of static CrowdHuman images. For the mixed training set of LQDanceTrack and DanceTrack, we performed joint training using the CrowdHuman training set and validation set. For the mixed training set of LQMOT and MOT, we conducted joint training using the CrowdHuman validation set.

\subsection{Performance Comparison with the State-of-the-Art Methods on LQDanceTrack and DanceTrack}
To demonstrate the effectiveness of our method in real-world low-quality video scenarios and its ability to maintain good performance in conventional scenarios. We trained VSE-MOT and other state-of-the-art methods on the mixed training set of LQDanceTrack and DanceTrack with a ratio of low-quality videos to high-quality videos of 2:1, and tested them on the validation sets of LQDanceTrack and DanceTrack. The obtained results are presented in Table \ref{tab:1} and \ref{tab:2}, where the top-performing outcomes are highlighted with boldface. As can be seen from Table \ref{tab:1}, the proposed VSE-MOT outweighs other methods on the LQDanceTrack validation set and achieves the best in all metrics. Compared with other state-of-the-art methods, it leads by 8\% to 20\% in each metric, demonstrating the effectiveness of the proposed method in real-world low-quality video scenarios. As can be seen from Table \ref{tab:2}, the proposed VSE-MOT outperforms other methods on the DanceTrack validation set. Compared with other state-of-the-art methods, it leads by 2\% to 20\% in each metric. It shows that after training on the mixed training set, the proposed method can maintain good performance in conventional scenarios.
\begin{table}[H]
\centering
\caption{Performance Comparison of Different Trackers on the LQDanceTrack Validation Set. ↑ Indicates Higher Is Better, the Best Results for Each Metric Are Highlighted in Bold. The Proposed VSE-MOT Outweighs Other Methods on the LQDanceTrack Validation Set and Achieves the Best in All Metrics, Demonstrating the Effectiveness of the Proposed Method in Real-world Low-quality Video Scenarios.}
\label{tab:1}
\resizebox{\columnwidth}{!}{%
\begin{tabular}{c@{\hspace{4pt}}|@{\hspace{4pt}}ccccc}
\toprule
Tracker & HOTA(\%)↑ & DetA(\%)↑ & AssA(\%)↑ & MOTA(\%)↑ & IDF1(\%)↑ \\
\midrule
CO-MOT \cite{58} & 42.0 & 54.4 & 32.9 & 53.3 & 43.0 \\
MeMOTR \cite{59} & 49.1 & 58.1 & 42.1 & 63.7 & 52.0 \\
MOTIP \cite{39} & 52.2 & 61.2 & 45.3 & 67.8 & 57.1 \\
Hybrid-SORT-ReID \cite{60} & 41.8 & 52.3 & 33.7 & 59.8 & 44.3 \\
Hybrid-SORT \cite{60} & 37.3 & 52.4 & 26.8 & 59.6 & 38.9 \\
MOTR \cite{34} & 39.9 & 51.8 & 31.3 & 46.3 & 41.4 \\
MOTRv2(baseline) \cite{11} & 51.4 & 67.3 & 39.6 & 65.2 & 53.2 \\
\midrule 
VSE-MOT(ours) & \textbf{59.9} & \textbf{73.7} & \textbf{48.9} & \textbf{84.1}& \textbf{62.3} \\
\bottomrule
\end{tabular}%
}
\end{table}

\begin{table}[H]
\centering
\caption{Performance Comparison of Different Trackers on the DanceTrack Validation Set. ↑ Indicates Higher Is Better, the Best Results for Each Metric Are Highlighted in Bold. The Proposed VSE-MOT Outperforms Other Methods
on the DanceTrack Validation Set. It Shows That after Training on the Mixed Training Set, the Proposed Method Can Maintain Good
Performance in Conventional Scenarios.}
\label{tab:2}
\resizebox{\columnwidth}{!}{%
\begin{tabular}{c|ccccc}
\toprule
Tracker & HOTA(\%)↑ & DetA(\%)↑ & AssA(\%)↑ & MOTA(\%)↑ & IDF1(\%)↑ \\
\midrule
CO-MOT \cite{58}& 46.7& 54.8& 40.5& 44.9& 47.3\\
MeMOTR \cite{59}&  59.3& 69.9& 50.7& 78.1& 62.7\\
MOTIP \cite{39}& 62.1& 73.5& 52.8& 83.8& \textbf{66.9}\\
Hybrid-SORT-ReID \cite{60}& 49.3& 56.5& 43.5& 65.3& 47.4\\
Hybrid-SORT \cite{60}& 44.5& 56.3& 35.6& 64.5& 43.2\\
MOTR \cite{34}& 48.4& 59.2& 39.8& 57.5& 50.2\\
MOTRv2(baseline) \cite{11}& 58.4& 69.9& 49.0& 80.2& 61.4\\
\hline
VSE-MOT(ours) &  \textbf{64.1}& \textbf{77.2}& \textbf{53.4}& \textbf{86.4}& 65.9\\
\bottomrule
\end{tabular}%
}
\end{table}

\subsection{Performance Comparison with the State-of-the-Art Methods on LQMOT and MOT}
We trained VSE-MOT and other state-of-the-art methods on the mixed training set of LQMOT and MOT with a ratio of low-quality videos to high-quality videos of 2:1, and tested them on the validation sets of LQMOT and MOT. The obtained results are presented in Table \ref{tab:3} and \ref{tab:4}. As can be seen from Table \ref{tab:3}, the proposed VSE-MOT outweighs other methods on the LQDanceTrack validation set. Compared with other state-of-the-art methods, it leads by 4\% to 15\% in HOTA, DetA, and AssA, demonstrating the effectiveness of the proposed method in real-world low-quality video scenarios. As can be seen from Table \ref{tab:4}, the proposed VSE-MOT outperforms other methods on the DanceTrack validation set. Compared with other state-of-the-art methods, it leads by 6\% to 19\% in HOTA, DetA, and AssA. It shows that after training on the mixed training set, the proposed method can maintain good performance in conventional scenarios.

\begin{table}[H]
\centering
\caption{Performance Comparison of Different Trackers on the LQMOT Validation Set. ↑ Indicates Higher Is Better, the Best Results for Each Metric Are Highlighted in Bold. The Proposed
VSE-MOT Outweighs Other Methods on the LQMOT
Validation Set, Demonstrating the Effectiveness of the Proposed Method in Real-world Low-quality Video Scenarios.}
 \label{tab:3}
\resizebox{\columnwidth}{!}{%
\begin{tabular}{c|ccccc}
\toprule
Tracker & HOTA(\%)↑ & DetA(\%)↑ & AssA(\%)↑ & MOTA(\%)↑ & IDF1(\%)↑ \\\midrule
MeMOTR \cite{59}& 44.9& 53.3& 38.2& 62.1& 56.5\\
MOTIP \cite{39}& 46.6& 55.6& 39.4& 65.9& 57.8\\
 ByteTrack \cite{21}& 51.2& 59.0& 44.8& \textbf{71.3}&62.3\\
Hybrid-SORT-ReID \cite{60}& 49.4& 58.4& 42.0& 70.9& 59.8\\
Hybrid-SORT \cite{60}& 47.1& 56.8& 39.7& 68.4& 58.2\\
MOTR \cite{34}& 43.3& 48.7& 38.9& 55.3& 53.5\\
MOTRv2(baseline) \cite{11}& 49.1& 57.5& 42.3& 60.2& 53.7\\
\hline
VSE-MOT(ours) & \textbf{58.5}& \textbf{62.3}& \textbf{54.9}& 68.2& \textbf{64.1}\\
\bottomrule
\end{tabular}%
}
\end{table}
\begin{table}[H]
\centering
\caption{Performance Comparison of Different Trackers on the MOT Validation Set. ↑ Indicates Higher Is Better, the Best Results for Each Metric Are Highlighted in Bold. The Proposed VSE-MOT Outperforms Other Methods
on the MOT Validation Set. It Shows That after Training on the Mixed Training Set, the Proposed Method Can Maintain Good
Performance in Conventional Scenarios.}
\label{tab:4}
\resizebox{\columnwidth}{!}{%
\begin{tabular}{c|ccccc}
\toprule
Tracker & HOTA(\%)↑ & DetA(\%)↑ & AssA(\%)↑ & MOTA(\%)↑ & IDF1(\%)↑ \\\midrule
MeMOTR \cite{59}&  52.6& 59.9& 46.5& 67.4& 65.1\\
MOTIP \cite{39}& 54.3& 60.2& 49.3& 70.3& 67.5\\
 ByteTrack \cite{21}& 58.7& 64.4& 53.7& \textbf{76.7}&71.5\\
Hybrid-SORT-ReID \cite{60}& 57.5& 64.2& 51.7& 76.6& 69.9\\
Hybrid-SORT \cite{60}& 55.7& 63.7& 49.1& 72.3& 67.8\\
MOTR \cite{34}& 48.2& 51.0& 45.8& 58.6& 59.6\\
MOTRv2(baseline) \cite{11}& 55.3& 61.3& 50.2& 66.4& 61.8\\
\hline
VSE-MOT(ours) & \textbf{67.8}& \textbf{69.6}& \textbf{66.0}& 73.2& \textbf{71.7}\\
\bottomrule
\end{tabular}%
}
\end{table}

\subsection{Ablation Study}
\subsubsection{Verification of Proposed Components}
In this section, we investigate several components of our approach, including the frozen CLIP Image Encoder, MOT-Adapter, and VSFM module. In the ablation study, we adopt the MOTRv2 model as the baseline model. The baseline model is represented as No.1 and No.5 on the LQDanceTrack and DanceTrack validation sets in Table \ref{tab:5}, respectively. Table \ref{tab:5} summarizes the effects of these components on the LQDanceTrack and DanceTrack validation sets, and the improvements are consistent across both datasets. It can be observed from Table \ref{tab:5} that integrating the proposed modules into the baseline gradually enhances overall performance.
\begin{table}[H]
\centering
\caption{Effectiveness of Proposed Components on the LQDanceTrack and DanceTrack Validation Sets. \checkmark Represents the Baseline Equipped with the Corresponding Component. It Can Be Observed That Integrating
the Proposed Modules into the Baseline Gradually Enhances
Overall Performance.}
\label{tab:5}
\resizebox{\columnwidth}{!}{%
\begin{tabular}{c|c|cccc|ccccc}
\toprule
No.&  \multicolumn{1}{c|}{Dataset}& Baseline&CLIP Image Encoder& MOT-Adapter& VSFM& HOTA(\%)↑& DetA(\%)↑& AssA(\%)↑& MOTA(\%)↑&IDF1(\%)↑\\
\midrule
1
&  & \checkmark&& & & 51.4& 67.3& 39.6& 65.2&53.2\\
2
&  LQDanceTrack& \checkmark&\checkmark& & & 53.3& 71.2& 40.0& 75.7&54.2\\
3
&  & \checkmark&\checkmark& \checkmark& & 55.7& 72.4& 43.1& 78.4&56.9\\
 4& & \checkmark& \checkmark& \checkmark& \checkmark& \textbf{59.9}& \textbf{73.7}& \textbf{48.9}& \textbf{84.1}& \textbf{62.3}\\
 \hline
 5& & \checkmark& & & & 58.4& 69.9& 49.0& 80.2&61.4\\
 6& DanceTrack& \checkmark& \checkmark& & & 58.5& 72.9& 47.2& 82.1&61.7\\
 7& & \checkmark& \checkmark& \checkmark& & 60.1& 74.3& 48.7& 83.3&62.6\\
8&  & \checkmark&\checkmark& \checkmark& \checkmark&  \textbf{64.1}& \textbf{77.2}& \textbf{53.4}& \textbf{86.4}& \textbf{65.9}\\
\bottomrule
\end{tabular}%
}
\end{table}
After adding the frozen CLIP Image Encoder to the baseline (No.2, No.6), No.2 shows a significant improvement in HOTA and MOTA scores, demonstrating that incorporating global visual semantic information features from images can lead to better association performance. No.6 exhibits better performance in MOTA and DetA compared to the baseline, proving that incorporating global visual semantic information features from images results in more robust target detection and tracking performance when trained on mixed datasets. No.3 and No.7 in Table \ref{tab:5} represent the models obtained by adding the MOT-Adapter to the No. 2 and No.6 models, respectively. No.3 and No.7 shows improved performance across all metrics compared to No. 2 and No.6, indicating that the MOT-Adapter can better extract image features for multi-object tracking tasks, further enhancing the algorithm's robustness. No.4 and No.8  in Table \ref{tab:5} represent the models obtained by adding the VSFM module to the No.3 and No.7 models, respectively. The results of No.4 and No.8 demonstrate that the proposed module can better integrate the query vector with global visual semantic information of images, improving the model's generalization ability and further achieving better target detection and tracking performance.

\subsubsection{Analysis of Fusion Module}
In this section, we investigate the VSFM module of our approach. To validate the effectiveness of the proposed VSFM module, we conducted experiments on the LQDanceTrack and DanceTrack validation sets using the VSFM module and several common feature fusion modules. In the ablation study, a baseline model with the frozen CLIP Image Encoder and MOT-Adapter modules was defined. The baseline model is represented as No.1 and No.6 on the LQDanceTrack and DanceTrack validation sets in Table \ref{tab:6}, respectively. Table \ref{tab:6} summarizes the impact of different fusion modules on the LQDanceTrack and DanceTrack validation sets. It can be observed from Table \ref{tab:6} that the proposed VSFM module can enhance the model's tracking performance and achieve better results than other fusion modules.
\begin{table}[H]
\centering
\caption{Comparison of Applying Different Fusion Modules on the LQDanceTrack and DanceTrack Validation Sets. It Can Be Observed That the Proposed
VSFM Module Can Enhance the Tracking Performance of Model
and Achieve Better Results Than Other Fusion Modules.}
\label{tab:6}
\resizebox{\columnwidth}{!}{%
\begin{tabular}{c|c|c|ccccc}
\toprule
 No.& Dataset&Fusion Module& HOTA(\%)↑ & DetA(\%)↑ & AssA(\%)↑ & MOTA(\%)↑ & IDF1(\%)↑ \\
\midrule
 1& &Baseline& 55.7& 72.4& 43.1& 78.4& 56.9\\
 2& &Additive Fusion& 56.6& 73.5& 43.7& 82.3& 57.6\\
 3& LQDanceTrack&Concatenation Fusion& 56.3& 72.3& 44.0& 81.1& 57.6\\
 4& &CBAM Fusion& 57.3& 72.8& 45.3& 81.7& 59.2\\
 5& &VSFM(ours)& \textbf{59.9}& \textbf{73.7}& \textbf{48.9}& \textbf{84.1}& \textbf{62.3}\\
\hline
 6& & Baseline& 60.1& 74.3& 48.7& 83.3&62.6\\
 7& & Additive Fusion& 61.4& 75.5& 50.1& 84.5&62.9\\
 8& DanceTrack& Concatenation Fusion& 61.2& 75.7& 49.7& 84.8&62.8\\
 9& & CBAM Fusion& 61.9& 76.1& 50.7& 85.0&63.4\\
 10& & VSFM(ours)&  \textbf{64.1}& \textbf{77.2}& \textbf{53.4}& \textbf{86.4}& \textbf{65.9}\\
 \bottomrule
\end{tabular}%
}
\end{table}
For the LQDanceTrack validation set, after adding Additive Fusion to the baseline (No.2), the HOTA score improved by about 1\%, and the MOTA score improved by about 4\%. After adding Concatenation Fusion to the baseline (No.3), the HOTA score improved by about 1\%, and the MOTA score improved by about 3\%. After adding 5-layer CBAM Fusion to the baseline (No.4), the HOTA score improved by about 2\%, and the MOTA score improved by about 3\%. After adding the VSFM module to the baseline (No.5), the HOTA score improved by about 4\%, and the MOTA score improved by about 7\%, indicating that the VSFM module can effectively fuse the query vector with global visual semantic information of images in real-world low-quality video scenarios, and the fusion effect is better than other common fusion methods.

For the DanceTrack validation set, after adding Additive Fusion to the baseline (No.7), the HOTA score improved by about 1\%, and the MOTA score improved by about 1\%. After adding Concatenation Fusion to the baseline (No.8), the HOTA score improved by about 1\%, and the MOTA score improved by about 1\%. After adding 5-layer CBAM Fusion to the baseline (No.9), the HOTA score improved by about 2\%, and the MOTA score improved by about 2\%. After adding the VSFM module to the baseline (No.10), the HOTA score improved by about 4\%, and the MOTA score improved by about 3\%, indicating that the VSFM module can also effectively fuse the query vector with global visual semantic information of images in real-world conventional scenarios, and the fusion effect is better than other common fusion methods.

\subsubsection{Analysis of Mixed Training Set}
In this section, we investigate the impact of the ratio of high-quality to low-quality data in the mixed training set on the algorithm's performance, and validate it on the validation sets of LQDanceTrack and DanceTrack in Table \ref{tab:7}. When trained exclusively on high-quality data, the proposed method exhibits suboptimal performance on the LQDanceTrack validation set. In contrast, it demonstrates strong performance on the DanceTrack validation set. At a 1:1 ratio of high-quality to low-quality data, the proposed method undergoes a substantial performance boost on the LQDanceTrack validation set, with the MOTA score and the HOTA score improving by approximately 8\%. On the DanceTrack validation set, there is a marginal decline of around 1\% in both metrics. When the ratio of high-quality to low-quality data is adjusted to 1:2, both metrics show slight improvements across both validation sets. Training solely on low-quality data leads to performance degradation for the proposed method on both the LQDanceTrack and DanceTrack validation sets, yielding results inferior to those obtained with the mixed training set. These experimental findings indicate that, in comparison to training on purely high-quality or purely low-quality data, training on a mixed dataset enables the model to achieve superior performance in both low-quality scenarios and regular scenarios. Furthermore, the model trained with a 1:2 ratio of high-quality to low-quality data delivers the best overall performance.
\begin{table}[H]
\centering
\caption{Comparison of Applying Different Ratio of High-quality to Low-quality Data in the Mixed Training Set  on the LQDanceTrack and DanceTrack Validation Sets. It Can Be Observed Better Results Are Achieved When the Ratio of High-quality Data to Low-quality Data Is 1:2.}
\label{tab:7}
\resizebox{\columnwidth}{!}{%
\begin{tabular}{c|c|c|ccccc}
\toprule
 No.& Dataset&Low-quality : High-quality& HOTA(\%)↑ & DetA(\%)↑ & AssA(\%)↑ & MOTA(\%)↑ & IDF1(\%)↑ \\
\midrule
 1& &All High-quality& 49.6& 64.4& 38.5& 74.4& 53.6\\
 3& LQDanceTrack&1:1& 57.4& \textbf{73.9}& 44.9& 82.6& 57.9\\
 4& &2:1& \textbf{59.9}& 73.7& \textbf{48.9}& \textbf{84.1}& \textbf{62.3}\\
 5& &All Low-quality& 56.1& 71.8& 43.8& 80.0& 56.1\\
\hline
 6& & All High-quality& 64.1& \textbf{77.8}& 52.9& \textbf{87.0}&65.7\\
 8& DanceTrack& 1:1& 63.4& 77.7& 52.0&  86.4&64.0\\
 9& & 2:1&   64.1& 77.2& \textbf{53.4}& 86.4& \textbf{65.9}\\
 10& & All Low-quality& 61.2&  77.0& 48.8&  86.5&63.1\\
 \bottomrule
\end{tabular}%
}
\end{table}

\subsubsection{Efficiency Analysis}
In this section, we investigate the impact of introduced components in this paper on the efficiency of the multi-object tracking using the LQDanceTrack validation set in Table \ref{tab:8}. The integration of the CLIP image encoder introduces a substantial number of additional parameters, which significantly degrades efficiency. In contrast, the MOT-Adapter and VSFM module involve only a small number of extra parameters and do not exert a noticeable impact on efficiency.
\begin{table}[H]
\centering
\caption{Efficiency Analysis on the Impact of Introduced Components on Multi-Object Tracking Using the LQDanceTrack Validation Set. It Can Be Observed That the CLIP Image Encoder Is the Primary Contributor to the Degradation in Efficiency.}
\label{tab:8}
\resizebox{\columnwidth}{!}{%
\begin{tabular}{c|cccc|cc}\toprule

No.& Baseline&CLIP Image Encoder& MOT-Adapter& VSFM& Parameters (M)& FPS\\\midrule

1
& \checkmark&& & & 41.9& 15.8\\
2
& \checkmark&\checkmark& & & 81.4& 9.2\\
3
& \checkmark&\checkmark& \checkmark& & 81.9& 9.1\\
 4& \checkmark& \checkmark& \checkmark& \checkmark& 83.3& 8.7\\ \bottomrule 
\end{tabular}%
}
\end{table}

\subsection{Visualization}
\begin{figure}[H]
    \centering
    \includegraphics[width=1\linewidth]{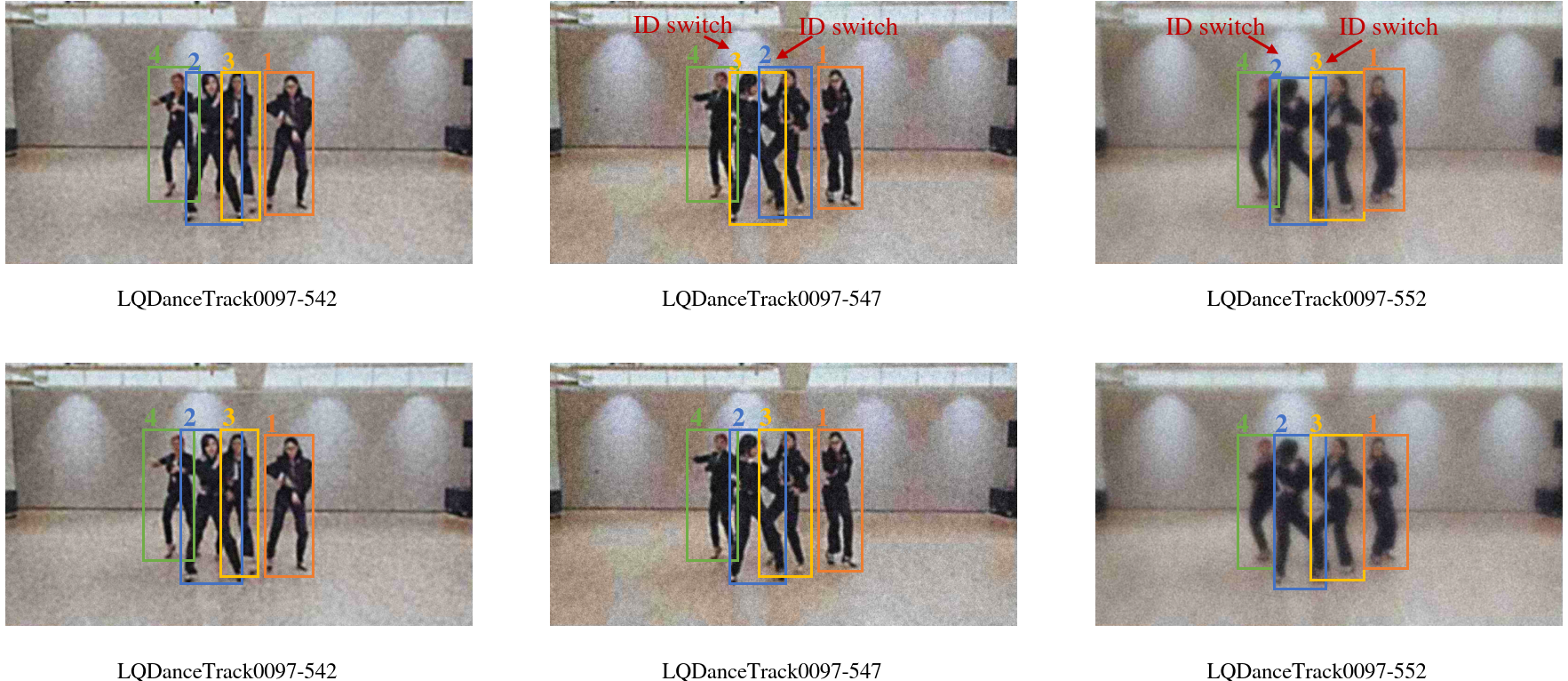}
    \caption{Partial results of the baseline method (first row) and VSE-MOT (second row) on the LQDanceTrack validation set. In low-quality videos, our method achieves better tracking performance than the baseline model by introducing global visual semantic information of the images.}
    \label{fig5}
\end{figure}

\begin{figure}[H]
    \centering
    \includegraphics[width=1\linewidth]{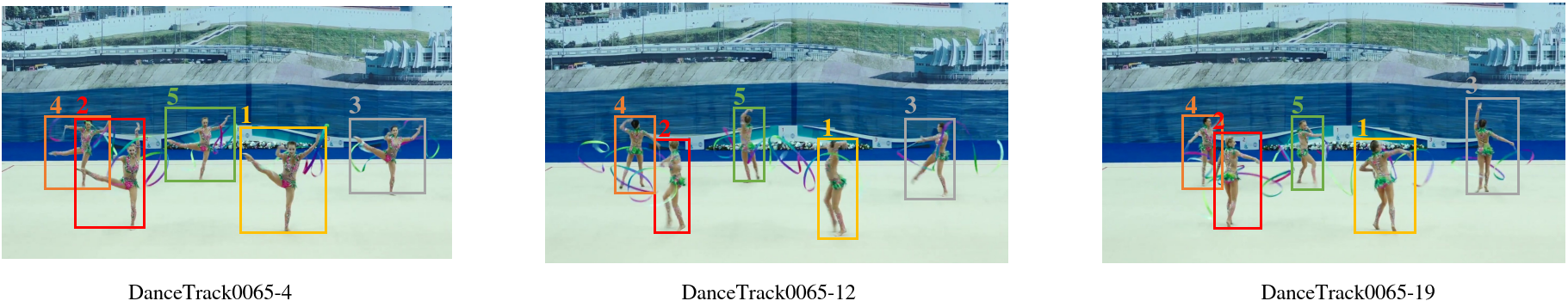}
    \caption{Partial results of VSE-MOT on the DanceTrack validation set. Our method
can maintain good performance in conventional scenarios.}
    \label{fig6}
\end{figure}

To visually demonstrate the effectiveness of VSE-MOT in low-quality videos, we selected some results on the low-quality video multi-object tracking dataset LQDanceTrack to show the performance. In this section, we adopt the MOTRv2 model as the baseline model. In Figure \ref{fig5}, the first and second rows depict the visual outcomes of the baseline method and our method, respectively. The distinct hues of the bounding boxes signify unique identities within each image. In the images LQDanceTrack0097-547 and LQDanceTrack0097-552 of the first row in Figure \ref{fig5}, No.2 and No.3 move rapidly, and the image quality is not very good, resulting in the phenomenon of incorrect exchange of identity (ID). Therefore, the frozen CLIP Image Encoder, MOT-Adapter, and VSFM modules are introduced in our method to handle low-quality video scenarios similar to LQDanceTrack0097-547 and LQDanceTrack0097-552. The second row of Figure \ref{fig5} shows that our method can maintain accurate tracking in low-quality videos.
To visually demonstrate the effectiveness of VSE-MOT in conventional scenarios, we selected some results from the DanceTrack dataset to show the performance. Figure \ref{fig6} corresponds to the visualization results of our method, and the colored bounding boxes denote distinct identities present in each image. Figure \ref{fig6} shows that our method can maintain good performance in conventional scenarios.
The visualizations mentioned above demonstrate the effectiveness and superiority of our proposed method in real-world low-quality video scenarios, while our method also maintains robust performance in conventional scenes, enhancing the robustness of multi-object tracking algorithms across different scenarios.

\section{Conclusion}
In this paper, an end-to-end online tracking framework named VSE-MOT is proposed for joint detection and tracking. To address the problem that the tracking performance of current multi-object tracking algorithms significantly degrades in low-quality video scenarios, the frozen CLIP Image Encoder is introduced to fully extract and explore the global visual semantic information of the images and fuse the query vectors and the global visual semantic information of the images, which can obtain more detailed target identity information. In addition, considering the problem that the original image features extracted by the frozen CLIP Image Encoder are not suitable for the multi-object tracking task, the MOT-Adapter module is introduced to provide more global visual semantic information of the images that is more in line with the multi-object tracking task. Finally, to better fuse the query vectors and the global visual semantic information of the images, we propose a feature fusion module named VSFM. Extensive experimental results demonstrate the effectiveness of each module. The proposed method is effective and superior in low-quality video scenarios in the real world and can maintain good performance in conventional scenarios. However, the introduction of the frozen CLIP Image Encoder increases many parameters of the model, affecting the efficiency of the model. Therefore, in future work, we will explore multiple strategies to optimize the efficiency of the model while maintaining its powerful functionality.




 \bibliographystyle{elsarticle-num} 
 \bibliography{VSE-MOT}






\end{document}